\documentclass[10pt,twocolumn,letterpaper]{article}

\usepackage{iccv}              

\definecolor{iccvblue}{rgb}{0.21,0.49,0.74}
\usepackage[pagebackref,breaklinks,colorlinks,allcolors=iccvblue]{hyperref}
\usepackage{utfsym}
\usepackage{multirow}
\usepackage{array}

\def\paperID{2490} 
\def\confName{ICCV}
\def\confYear{2025}

\title{Deepfake Detection via Knowledge Injection}

\author{Tonghui Li \,\, Yuanfang Guo\,\, Heqi Peng  \,\, Zeming Liu \,\, Yunhong Wang \\
School of Computer Science and Engineering, Beihang University, China \\
{\tt\small {lthlth, AndyGuo, penghq, zmliu, yhwang}@buaa.edu.cn} \\
}

\begin{document}
\maketitle
\begin{abstract}

Deepfake detection technologies become vital because current generative AI models can generate realistic deepfakes, which may be utilized in malicious purposes. Existing deepfake detection methods either rely on developing classification methods to better fit the distributions of the training data, or exploiting forgery synthesis mechanisms to learn a more comprehensive forgery distribution. Unfortunately, these methods tend to overlook the essential role of real data knowledge, which limits their generalization ability in processing the unseen real and fake data. To tackle these challenges, in this paper, we propose a simple and novel approach, named Knowledge Injection based deepfake Detection (KID), by constructing a multi-task learning based knowledge injection framework, which can be easily plugged into existing ViT-based backbone models, including foundation models. Specifically, a knowledge injection module is proposed to learn and inject necessary knowledge into the backbone model, to achieve a more accurate modeling of the distributions of real and fake data. A coarse-grained forgery localization branch is constructed to learn the forgery locations in a multi-task learning manner, to enrich the learned forgery knowledge for the knowledge injection module. Two layer-wise suppression and contrast losses are proposed to emphasize the knowledge of real data in the knowledge injection module, to further balance the portions of the real and fake knowledge. Extensive experiments have demonstrated that our KID possesses excellent compatibility with different scales of Vit-based backbone models, and achieves state-of-the-art generalization performance while enhancing the training convergence speed.

\end{abstract}    
\section{Introduction}
\label{sec:intro}

The rapid advancement of generative AI models, such as GANs  \cite{goodfellow_generative_2014} and VAEs  \cite{kingma2013auto}, have significantly promoted the development of deepfake technology. Since highly realistic deepfakes (a.k.a. deep facial forgeries) can be exploited for malicious purposes, such as financial fraud  \cite{2024financial} or political defamation  \cite{2022Zelensky}, which tends to induce severe consequences, deepfake detection becomes vital in fighting against these deepfakes. With the growing prevalence of deepfakes on social media platforms, it is imperative to develop deepfake detection methods with high generalization ability to handle the deepfakes produced by unseen generative methods.

\begin{figure}[t]
  \centering
   \includegraphics[width=1.9\linewidth]{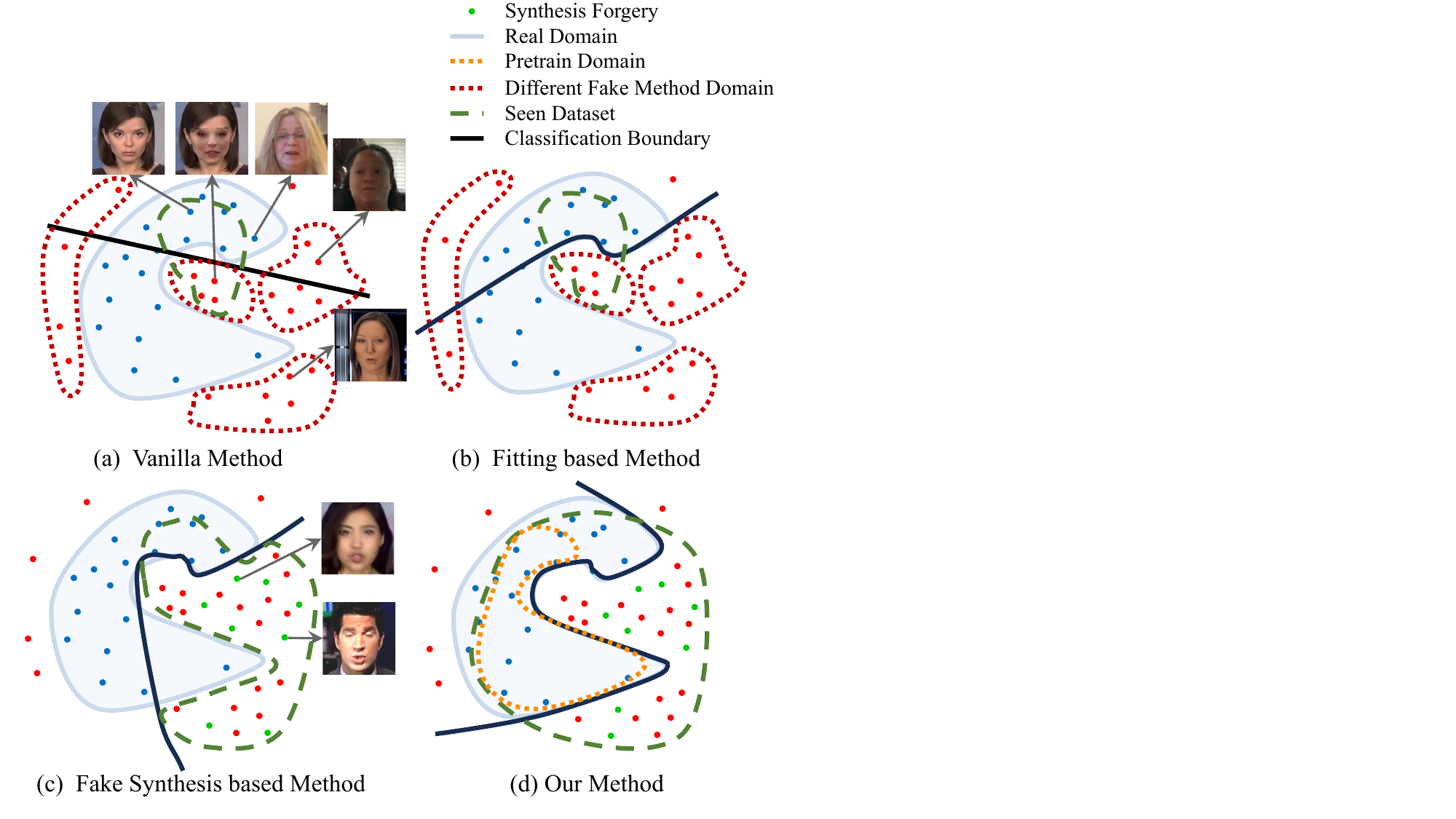}

   \caption{Biased classification boundary caused by the model's insufficient comprehension of real or fake image distribution.
   The vanilla method and fitting based method are usually limited to the training set, resulting in a biased classification boundary. Fake synthesis based methods have a better understanding of fake image distribution and thus establish more effective boundaries, but still lack a robust grasp of the characteristics of real images. Our proposed approach achieves a more thorough understanding of both real and fake images.}
   \label{fig:Motivation}
\end{figure}

The majority of the previous methods  \cite{afchar2018mesonet,rossler2019faceforensics++,li2018exposing,liu2020global,zhao2021learning,bai2023aunet} mainly rely on the learning capacity of deep neural networks (DNNs) to learn certain forgery cues from the training set, to differentiate deepfakes from real data. Specifically, early vanilla methods  \cite{afchar2018mesonet,rossler2019faceforensics++}, which only utilize DNNs, directly learn the detection cues from the real and fake data, as shown in \cref{fig:Motivation}a. Then, the fitting based methods  \cite{dang2020detection,zhao2021multi,liu2021spatial}, which improve the vanilla method by exploiting special designs such as attention  mechanism, further refine the decision boundary to better fit the distributions of the training real and fake data, as shown in \cref{fig:Motivation}b. Unfortunately, these methods tend to perform decently within the domain of the training data while struggling with the testing data from unseen forgery techniques. Meanwhile, fake synthesis based approaches  \cite{li2020face, shiohara2022detecting} generate fake training data via forgery synthesis pipelines based on prior knowledge, e.g. blending boundaries, and establish more effective decision boundaries, as shown in \cref{fig:Motivation}c. Although these methods enable the backbone DNNs to learn a more comprehensive forgery distribution, they have not paid enough attentions to the distribution of real images. Their understanding of the real data is only limited to the training data, which makes them difficult to process the unseen real data. Therefore, these methods can hardly establish an accurate classification boundary, which is simultaneously correlated to the distributions of both the real and deepfake data, to differentiate the unseen deepfakes from the real data.

To overcome the aforementioned issues, in this paper, we propose a novel approach, named Knowledge Injection based deepfake Detection (KID). Specifically, we propose a multi-task learning based Knowledge Injection framework, which contains a knowledge injection module, a coarse-grained forgery localization branch and two layer-wise suppression and contrast losses (S\&C). To preserve the real data knowledge in the pre-trained backbone model obtained from the pre-training stage while integrating the understanding of deepfake data, we propose a knowledge injection module to progressively learn the distributions of real and fake data, and inject the learned knowledge into the backbone model. To constrain the knowledge injection module to learn more forgery cues and enhance the location-awareness of injected forgery knowledge, we construct a coarse-grained forgery localization branch by capturing the consistency across regions in real images and the inconsistencies within fake images. To emphasize the knowledge of real data to balance the portions of real and fake knowledge, we propose two layer-wise suppression and contrast losses. The entire framework also utilizes self-blended images  \cite{chen2022self} as the training fake data to boost the model's capacity to generalize across unseen deepfake techniques. In general, the framework can simultaneously learn the distributions of real and fake data and learn more generalized fake features, to establish a more accurate classification boundary for unseen real and fake data.  
    
Our major contributions are summarized as follows:

\begin{itemize}
    \item 
    We propose a novel deepfake detection method, named Knowledge Injection based deepfake Detection (KID), by constructing a multi-task learning based knowledge injection framework, which can be easily plugged into existing ViT-based models, including foundation models.
    \item 
    We propose a knowledge injection module to preserve the knowledge of real data while injecting the understanding of deepfake data into the backbone model, to enable a more accurate modeling of the distributions of real and fake data.
    \item 
    We construct a coarse-grained forgery localization branch to enrich the learned forgery knowledge, including the forgery locations, in a multi-task learning manner.
    \item 
    We propose two layer-wise suppression and contrast losses to constrain the knowledge injection module, to emphasize the knowledge of real data to further balance the portions of real and fake knowledge.
    \item
    Extensive experiments have demonstrated that Our KID possesses excellent compatibility with different scales of Vit-based backbone models, and achieves state-of-the-art generalization performance while enhancing the training convergence speed.
\end{itemize}

\section{Related Work}
\label{sec:formatting}

In recent years, deepfake detection techniques have been continuously evolving. Early biometric based methods primarily concentrated on heuristic features, such as head pose \cite{yang2019exposingheadpose}, eye blinking \cite{li2018ictu}, and optical flow \cite{amerini2019deepfake}, etc. However, these approaches usually struggle to remain effective when processing high-quality deepfake data. 

Meanwhile, early vanilla methods  \cite{afchar2018mesonet,rossler2019faceforensics++}, which only utilize DNNs, directly learned the detection cues from the real and fake data, and drew massive attentions from researchers. Then, researchers quickly started to focus on improving the learning ability of DNNs, and numerous subsequent methods have been developed. From the perspective of training mechanism, these methods can be primarily classified into two categories, fitting based methods \cite{afchar2018mesonet,rossler2019faceforensics++,zhao2021multi,qian2020thinking} and fake synthesis based methods \cite{li2020face,dong2022protecting,chen2022self,shiohara2022detecting}.

\subsection{Fitting based Method}

Among the fitting based methods, some of them  \cite{dang2020detection,zhao2021multi} introduce attention mechanism to highlight suspicious regions in fake images. Meanwhile, methods like SPSL  \cite{liu2021spatial} and F3-Net  \cite{qian2020thinking} explore the frequency domain to identify forged content, which is not obvious in the spatial domain. Other techniques, such as PCL  \cite{zhao2021learning} and PD  \cite{zhang2022patch}, compare real and fake segments within the image to reveal inconsistencies. Despite these advances, these fitting-based methods primarily focus on differentiating real and fake data with cues learned from the training data, which makes them less effective against unseen data. 

\begin{figure*}
    \centering
    \includegraphics[width=\textwidth]{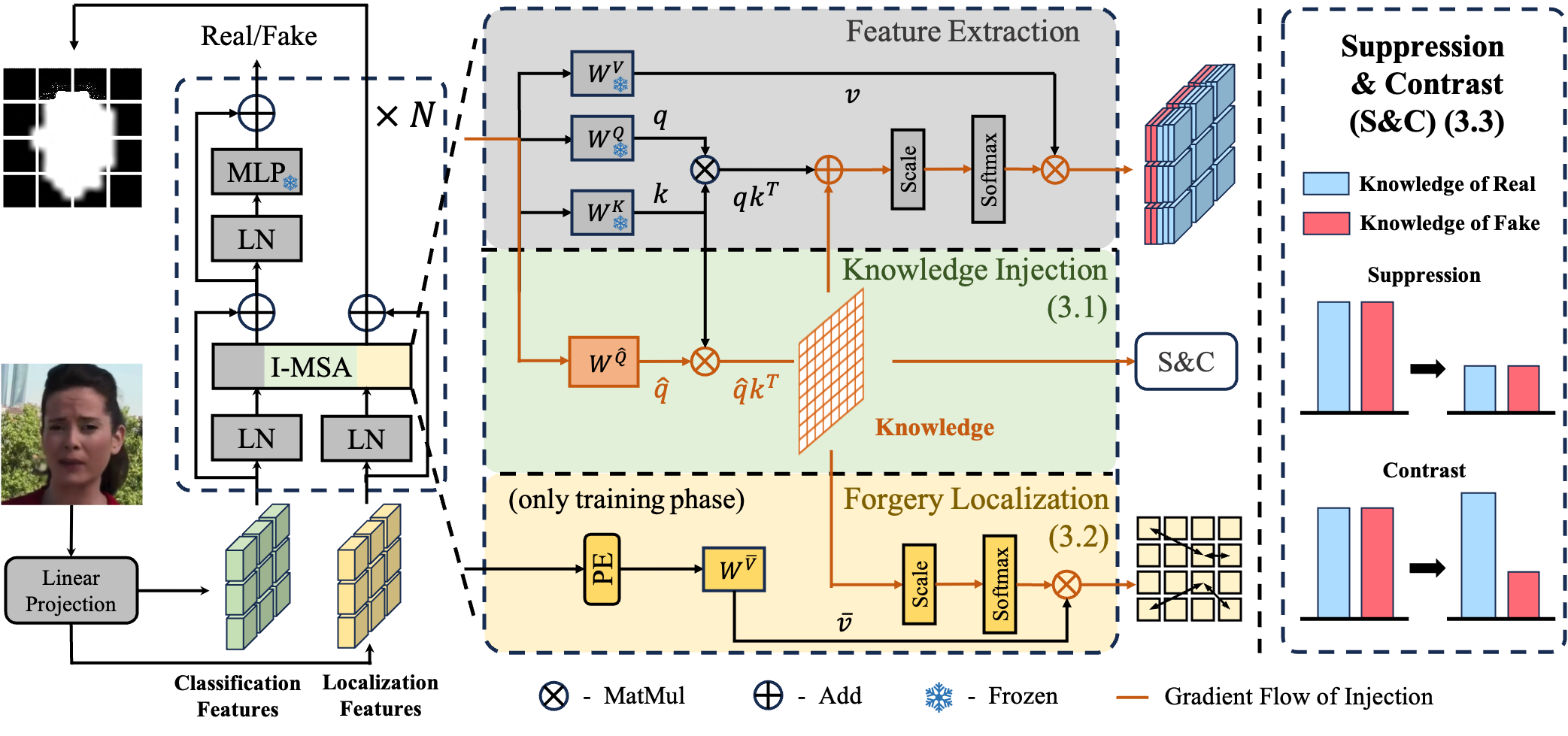}
    \caption{Overview of the Knowledge Injection based deepfake Detection framework. }
    \label{fig:Framework}
\end{figure*}

\subsection{Fake Synthesis based Method}

By recognizing the limitations of fitting based methods, researchers also explored generating fake images via forgery synthesis pipelines, to help the detection model to learn a more general knowledge from the fake training data. Then, the fake synthesis based methods became less overfitted to the training data and deepfake techniques. Face X-ray \cite{li2020face} firstly proposed to swap faces with similar facial landmarks and enable the detection model to focus on the blending boundaries. Then, ICT \cite{dong2022protecting} exploited this mechanism to help its identity learning and SLADD \cite{chen2022self} provided more diverse editing content. SBI \cite{shiohara2022detecting} further improved the forgery synthesis mechanisms by creating more realistic fake images with diverse forgery types, in a self-blending manner. These methods can learn a more general distribution of the fake data and establish a more effective decision boundary.

While these fake synthesis methods focus on improving the generalization ability of handling unseen fake data, they usually overlook the importance of the real data distributions, which makes them difficult to process the unseen real data. To address this issue, based on the strengths of fake data synthesis, we further capture a more precise representation of real and fake data distributions to better handle unseen data, via our proposed knowledge injection framework, which is compatible with ViT-based DNN models (including foundation models). With our proposed method, we provide a new direction for handling emerging deepfakes and enhancing the generalization ability to fight against the unseen deepfake techniques.

\section{Methodology}

Existing forgery detection methods usually overlook the role of real images, leading to imprecise classification boundaries and making it difficult to distinguish unseen real and fake data. To address these challenges, we propose the Knowledge Injection based deepfake Detection (KID) method by constructing a multi-task learning based knowledge injection framework. Our framework comprises three main components: a knowledge injection module, a coarse-grained forgery localization branch, and two layer-wise suppression and contrast losses, as illustrated in \cref{fig:Framework}. By leveraging these components along with a ViT-based backbone model and the self-blended training fake data, our KID can effectively learn the distributions of both real and fake images and establish a more precise classification boundary to distinguish the unseen real and fake images. Note that in this section we employ ViT \cite{dosovitskiy2020image} as an example backbone model for convenience.

\subsection{Knowledge Injection Module}

Since the employed ViT-based pre-trained model has been trained on the pre-training datasets such as ImageNet, which only contains real images, the backbone model naturally possesses extensive knowledge about the distribution of real images. This motivates us to propose our knowledge injection module, to preserve the knowledge of real data in the pre-trained backbone model while injecting the understandings of deepfake data. 

The ViT-based pre-trained models have shown remarkable performances in many classification tasks including deepfake detection, which can be attributed to their core attention mechanism. To integrate our knowledge injection framework into the backbone model, we propose an Injection Multi-Head Self-Attention (I-MSA) block to replace the regular multi-head self-attention block to perform knowledge injection. 

Let $P_l$ represent the input features of the I-MSA block at the $l$-th layer. $P_l$ is firstly splitted into multi-head features, denoted as $H^{i}_{l}$. In addition to the standard QKV (Query, Key, Value) calculations, we introduce an additional knowledge query vector, $\overline{Q}$, to calculate the authenticity correlation between image patches as

\begin{equation}
    \overline{Q}=H^i_lW^{\overline{Q}}_l.
\end{equation}

The authenticity correlation matrix $\overline{Corr_l}$, which represents the learned knowledge about the distributions of real and fake data, is then calculated based on $\overline{Q}$ and the key vector $K$, as

\begin{equation}
    \overline{Corr_l^i} = \frac{\overline{Q_l^i}K_l^i}{\sqrt{d_k}}.
\end{equation}

After the computation of the detection-related knowledge, it is further injected into the features obtained from the standard QKV calculations. Then, the self-attention is calculated as \cref{eq:self-attention}, by utilizing the injected features to complete the feature update

\begin{equation}\label{eq:self-attention}
    head_l^i = {\rm softmax}(\frac{Q_l^iK_l^i}{\sqrt{d_k}} + \overline{Corr_l^i})V^i_l.
\end{equation}

Via the adoption of the knowledge injection module, our KID can preserve the knowledge of real data learned by the backbone network, and obtain a more general representation of the real data distribution. Meanwhile, the module also acquire the knowledge of fake data distribution with the help of our coarse-grained forgery localization branch. Constrained by our layer-wise suppression and contrast losses, the knowledge injection module can acquire a more general knowledge of the authenticity-related cues and inject this knowledge into the backbone model to improve the modeling of both the real and fake data.

In the training process, the original attention branch is frozen and only the weight matrix $W^{\overline{Q}}_l$ is updated. The gradient flow in the I-MSA block is denoted by the orange-colored path in \cref{fig:Framework}. Under such circumstance, our framework can significantly reduce the amount of parameters to be updated, which simplifies the learning of the knowledge injection module, to achieve a faster convergence speed compared to the full-parameter tuning.

\subsection{Coarse-Grained Forgery Localization Branch}

To guide the knowledge injection module in learning the knowledge of fake data and enhance the location-awareness of the injected knowledge, we construct a separate coarse-grained forgery localization branch, as shown in \cref{fig:Framework}. Different from the original classification branch, our localization branch provides a supplementary pathway to constrain the detection model in a multi-task learning manner.

Before the processing of the transformer blocks, the initial localization features are obtained via identical approach as the initial classification features but with separate parameters. Once the auxiliary localization features $L_l$ are fed into the I-MSA block, Positional Encoding(PE) is applied to enhance the perception of positional information. 

During the update process, the localization features $L_{l+1}$ is finally updated based on the knowledge authenticity correlation matrix, as
\begin{equation}
    L_{l+1} = {\rm softmax} (\overline{Corr_l}) \cdot {\rm LN}(L_l+PE) \cdot 
 W_l^{\overline{K}}.
\end{equation}

Since the primary goal of our method is deepfake detection rather than accurate forgery localization, no upsampling operation is performed at the output of the entire network to avoid additional computations. Instead, a MLP layer is constructed to conduct coarse-grained classifications on the output image patch features of the last layer.

The corresponding coarse-grained localization ground truth is calculated based on the percentage $\Gamma_i$ of the number of pixels belonging to the outer face within each patch. The label of the i-th patch $y_i$ is defined as
\begin{equation}
    y_i = 
    \begin{cases}
    0, & \Gamma_i < \gamma_0\\
    1, & \Gamma_i > \gamma_1 \\
    \Gamma_i , & {\rm otherwise}
    \end{cases},
    \label{eq:coarse patch label}
\end{equation}
where $\gamma_0$ and $\gamma_1$ are hyperparameters which stand for the thresholds. For each patch, the percentage of the number of pixels belonging to the outer face $\Gamma_i$ within each patch is compared to these thresholds to label the patch as either an inner face patch or an outer face patch. Otherwise, the label is directly assigned based on the value of $\Gamma_i$, which represents the boundary region. 

At last, a dice loss is calculated to update the coarse-grained forgery localization branch. With the coarse-grained forgery localization branch, the knowledge injection module can better perceive spatial information and characterize the distribution of fake data.

\noindent\textbf{Remark.}
\cref{eq:self-attention} can be reformed to
\begin{equation}
    head_l^i = {\rm softmax}(\frac{H_l^i(W_l^Q+W_l^{\overline{Q}})K_l^i}{\sqrt{d_k}})V^i_l.
\end{equation}
Apparently, $W_l^Q$ and $W_l^{\overline{Q}}$ are symmetric. Then, 
\begin{equation}
    \frac{\partial L_{CE}}{\partial W_l^Q}=\frac{\partial L_{CE}}{\partial W_l^{\overline{Q}}},
\end{equation}
where $L_{CE}$ represents the cross-entropy loss. 
This observation implies that if the backbone network only incorporates the knowledge injection module, its training process is actually equivalent to updating the parameter $W_l^Q$ of a ViT model to fit the training set. Then, the detection model behaves similarly to a fitting-based method. By incorporating the coarse-grained forgery localization branch, we can disrupt this symmetry and constrain the knowledge injection module to concentrate on learning the forgery cues.

\subsection{Layer-Wise Suppression and Contrast Losses}
\label{subsec:Layer-Wise Contrast and Suppression}

With the help of the coarse-grained forgery localization branch, the detection model can concentrate on learning the forgery cues. However, without a proper constraint for the knowledge learning process, the detection model may over-emphasize the knowledge of fake data and can hardly balance the learned knowledge of the real and fake data. Under such circumstance, the detection model may still overfit to certain specific characteristics of the training set and then hurt the generalization ability.

To reasonably emphasize the knowledge of real data within the knowledge injection module and further balance the contributions of real and fake knowledge, we propose two layer-wise suppression and contrast losses. Unlike the coarse-grained forgery localization branch, our layer-wise suppression and contrast losses provide fine-grained control across each layer, to properly constrain the knowledge learned by the knowledge injection module.

Specifically, we suppress the activation values in the authenticity correlation matrix at the shallow layers, as described in \cref{eq:activation value} and \cref{eq:loss shallow}. This suppression loss constrains the detection model to maintain the core understandings of real data distributions at the shallow layers

\begin{equation}
    A_l = \frac{1}{MN}\sum_{i=0}^{M}\sum_{j=0}^{N}|\overline{Corr_{l,i,j}}|,
    \label{eq:activation value}
\end{equation}

\begin{equation}
    L_{S} = \sum_{l=0}^{L_0}\frac{\sum_{b=0}^{B}{\rm max}(0,\ A_l-\beta)}{B}.
    \label{eq:loss shallow}
\end{equation}

Note that $A_l$ calculates the averaged activation value of the authenticity correlation matrix $\overline{Corr_l}$, which measures the extent of knowledge (modifications) to be injected to the features extracted from the backbone model. Meanwhile, to allow the knowledge injection module to still learn the necessary knowledge for effective deepfake detection, we introduce an upper limit parameter $\beta$ to prevent excessive suppression. Then, the suppression loss $L_S$ is computed as the sum of losses across layers $0-L_0$.

At deep layers, the suppression loss will not be utilized and the contrast loss will be applied. According to general observations, real images generally exhibit strong internal consistencies and correlations across patches, whereas fake images often display inconsistencies between manipulated and benign regions, due to the imperfections of deepfake techniques. Therefore, we formulate the contrast loss based on the above intuitions, as 
\begin{equation}
    L_{D}= \\ 
    \sum_{l=L -2}^{L}\frac{\sum_{b=0}^{B}{\rm max}(0,\  A_l^{fake}-A_l^{real}+\mu)}{B},
    \label{eq:loss deep}
\end{equation}
where $A_l^{real}$ and $A_l^{fake}$ represent the activation values for the deep layers corresponding to the real and fake images, respectively. Note that $\mu$ is a hyperparameter that sets the minimum acceptable difference between the activation values of real and fake images. During the training process, the contrast loss is computed based on the authenticity correlation matrices (knowledge) from the I-MSA blocks in the final two transformer blocks. By applying the contrast loss, the detection model effectively improves the knowledge of the distributions of real and fake images, and thus establishes a more accurate classification boundary.

\subsection{Overall Loss}

At last, the overall loss function of our KID, as shown in \cref{eq:final loss}, is formed by integrating the cross entropy loss of the deepfake classification, the dice loss for coarse-grained forgery localization branch, and the layer-wise suppression and contrast losses. Note that the entire coarse-grained forgery localization branch is only utilized in the training phase.

\begin{equation}
    L=L_{CE} + L_{DICE} + L_{S} + L_{D}
    \label{eq:final loss}
\end{equation}

\section{Experiments}
\subsection{Experimental Settings}
\noindent\textbf{Datasets.} Following the settings used in  \cite{chen2022self}, we adopt the FF++ \cite{rossler2019faceforensics++} as the training set. The dataset contains 1,000 real images and 4,000 fake images generated by four manipulation methods: Deepfakes (DF) \cite{2017deepfakes}, Face2Face (F2F) \cite{thies2016face2face}, FaceSwap (FS) \cite{2016faceswap}, and NeuralTextures (NT) \cite{thies2019deferred}. To assess the generalization ability and stability of the training model across different datasets, we also conduct tests on four additional datasets: Celeb-DF-v2(CDF) \cite{li2020celeb}, DeepFakeDetection(DFD) \cite{google2019DFD}, DeepFake Detection Challenge public test set(DFDC) \cite{google2019DFDC} and Wild-Deepfake(FFIW) \cite{zhou2021face}. These alternative datasets contain a greater number and higher quality of fake images, providing a robust platform for evaluating the model's performance.\vspace{5pt}

\noindent\textbf{Evaluation Metrics.} To thoroughly verify the generalization ability of our method and compare its performance with previous methods, we adopt the evaluation protocol from  \cite{chen2022self}, we utilize the area under the receiver operating characteristic curve(AUC) as the primary evaluation metric, following  \cite{chen2022self}. \vspace{5pt}

\begin{table*}
\centering
\begin{tabular}{lp{1.8cm}<{\centering}p{0.8cm}<{\centering}p{0.8cm}<{\centering}p{1.2cm}<{\centering}p{1.2cm}<{\centering}p{1.2cm}<{\centering}p{1.2cm}<{\centering}}
\toprule
\multirow{2}{*}{Methods} & \multirow{2}{*}{Input Type} & \multicolumn{2}{c}{Training Set} & \multicolumn{4}{c}{Test Set AUC(\%)} \\
\cmidrule{3-8}
& & Real & Fake & DFD & CDF & DFDC & FFIW \\
\midrule
LipForensics \cite{haliassos2021lips} & Video & \usym{2713} & \usym{2713} & - & 82.40 & 73.50 & -\\

FTCN \cite{zheng2021exploring} & Video & \usym{2713} & \usym{2713} & 94.40 & 86.90 & 71.00 & 74.47 \\

FInfer \cite{hu2022finfer} & Video & \usym{2713} & \usym{2713} & - & 70.60 & - & 69.46 \\

FADE \cite{tan2023deepfake} & Video & \usym{2713} & \usym{2713} & 96.23 & 77.46 & - & - \\

AltFreezing \cite{wang2023altfreezing} & Video & \usym{2713} & \usym{2713} & 98.50 & 89.50 & - & - \\

\midrule

ViT \cite{dosovitskiy2020image} & Frame & \usym{2713} & \usym{2713} & 89.75 & 73.49 & 69.90 & - \\

Face X-ray \cite{li2020face} & Frame & \usym{2713} & \usym{2713} & 93.47 & - & - & - \\

PCL+I2G \cite{zhao2021learning} & Frame & \usym{2713} & \usym{2717} & 99.07 & 90.03 & 67.52 & - \\

ICT \cite{dong2022protecting} & Frame & \usym{2713} & \usym{2717} & 84.13 & 85.71 & - & - \\

DCL \cite{sun2022dual} & Frame & \usym{2713} & \usym{2717} & 91.66 & 82.30 & - & 71.14 \\

SBI+Xception \cite{shiohara2022detecting} & Frame & \usym{2713} & \usym{2717} & 97.56 & 93.18 & 72.42 & 76.72 \\

AUNet \cite{bai2023aunet} & Frame & \usym{2713} & \usym{2717} & \underline{99.22} & 92.77 & \underline{73.82} & \underline{81.45} \\

IID \cite{huang2023implicit} & Frame & \usym{2713} & \usym{2713} & 93.92 & 83.80 & - & - \\

IIL \cite{dong2023implicit} & Frame & \usym{2713} & \usym{2713} & 99.03 & 93.88 & - & - \\

LAA-Net \cite{nguyen2024laa} & Frame & \usym{2713} & \usym{2717} & 98.43 & \underline{95.40} & - &  80.03 \\

\midrule
KID (Ours) & Frame & \usym{2713} & \usym{2717} & \textbf{99.46} & \textbf{95.74} & \textbf{75.77} & \textbf{82.53} \\
\bottomrule

\end{tabular}
\caption{\textbf{Cross-dataset} evaluation in terms of AUC on DFD, CDF, DFDC and FFIW. The results of prior methods are directly cited from the original paper for a fair comparison. Bold and underlined values correspond to the best and the second-best values, respectively. Our method outperforms the state-of-the-art methods on all four datasets.}
\label{table:Cross-Dataset Evaluation}
\vspace{-5pt}
\end{table*}

\vspace{-5pt}
\noindent\textbf{Implementation details.} Throughout the experiment, RetinaFace \cite{deng2019retinaface} is used to detect the facial region, which is then cropped, and aligned, and the images are saved as input with the size of $224\times 224$. We adopt ViT/B-16 \cite{dosovitskiy2020image} pre-trained on ImageNet \cite{deng2009imagenet} as the model backbone.
In the training stage, the model is trained for a maximum of 300 epochs using the AdamW optimizer \cite{loshchilov2017decoupled}, with a weight decay of 0.01 and a batch size of 24. Early stopping is implemented, terminating training when the loss doesn't decrease for 20 consecutive epochs. The initial learning rate is set to $1\times 10^{-4}$, and we utilize cosine annealing for learning rate decay, with a lower bound of $1\times 10^{-6}$. The only updated parameters of the backbone part are class token and normalization layers. For data augmentation, we apply horizontal flipping, random hue saturation changes, random brightness contrast changes, JPEG compression, blurring and SBI \cite{chen2022self} fake synthesis. The upper and lower bounds in \cref{eq:coarse patch label} are defined as $\gamma_0=0.2$ and $\gamma_1=0.8$. Additionally, the boundary parameters in \cref{eq:loss shallow} and \cref{eq:loss deep} are set to $\beta=1.2$ and $\mu=0.1$. In the inference stage, we randomly extract 32 frames from each video for detection and take the average of the frame-level results as the video-level result. All experiments are conducted on four RTX 3080 Ti GPUs.

\begin{table}
    \centering
    \begin{tabular}{p{2.5cm}|p{1.5cm}<{\centering}p{1.5cm}<{\centering}}
    \toprule
        \multirow{2}{*}{Methods} &  \multicolumn{2}{c}{Test Set AUC(\%)}  \\ \cmidrule{2-3}
         & FF++ & CDF \\ 
         \midrule
        Xception \cite{chollet2017xception} & 99.70 & 65.30 \\
        EN-B4 \cite{tan2019efficientnet} & 99.22 & 66.24  \\
        MAT \cite{zhao2021multi}  & \textbf{99.80} & 76.65 \\ 
        $F^3$-Net \cite{qian2020thinking} & 98.10 & 65.17 \\
        Face X-ray \cite{li2020face} & 98.52 & 74.20 \\ 
        SPSL \cite{liu2021spatial} & 96.91 & 76.88 \\ 
        DCL \cite{sun2022dual} & 99.30 & 81.00 \\ 
        SLADD \cite{chen2022self} & 98.40 & 79.70 \\
        IID \cite{huang2023implicit} & 99.32 & 82.04 \\
        GS \cite{guo2023controllable} & 99.95 & 84.97 \\ 
        \midrule
        KID (Ours) & 98.91 & \textbf{86.32}  \\ \bottomrule
    \end{tabular}
    
    \caption{\textbf{Cross-dataset} evaluation for single-frame images without context on the CDF dataset by training on FF++.}
  \label{table:classification-2}
  \vspace{-4pt}
\end{table}



\subsection{Cross-Dataset Evaluation}
To demonstrate the superiority of our method in terms of generalizability, we evaluate the performance of our model, trained on the FF++ dataset, across various datasets and compare it to previous methods. The results are shown in \cref{table:Cross-Dataset Evaluation}.
\vspace{5pt}

\noindent\textbf{Comparison with Frame-Level Methods.} We first compare our method with the frame-level detection methods. Our method achieves state-of-the-art performance on all four datasets, outperforming the second-best result by 1.86\%, 1.95\% and 1.08\% on the high-quality forgery datasets Celeb-DF (CDF), DFDC and FFIW, respectively.  The results indicate that our proposed method has superior generalization and detection capabilities when faced with unknown, high-quality fake data. Additionally, certain methods only detect single-frame images without providing video-level results. For comparison with some frame-level methods, \cref{table:classification-2} shows the performance of our method against existing single-frame detection techniques on the CDF dataset. While our approach is slightly outperformed by MAT on the FF++ dataset, likely because it does not utilize fake images from FF++, it achieves the highest performance on CDF. This result demonstrates that our method performs at a cutting-edge level for single-image detection, even without relying on contextual information.
\vspace{5pt}

\noindent\textbf{Comparison with Video-Level Methods.} We then compare our method with various video-level detection methods. These methods typically require consecutive frames as input and leverage temporal correlation for detection. As shown in \cref{table:Cross-Dataset Evaluation}, even when randomly selecting frames, our method achieves the best performance, surpassing the state-of-the-art by 6.24\% and 4.77\% on the CDF and DFDC datasets, respectively. These results confirm that our method retains strong generalization performance without relying on temporal information.

\subsection{Cross-Manipulation Evaluation}
In deepfake detection tasks, not only the original images and post-forgery processing are critical, but the forgery technique itself also has a significant impact on forgery outcomes. To evaluate the model’s robustness across different methods, we tested the detection performance of our proposed method on various manipulation techniques.

\begin{table}
\begin{tabular}{lccccc}
\toprule
\multirow{2}{*}{Method} & \multicolumn{5}{c}{Test Set AUC(\%)}    \\ \cmidrule{2-6} 
& DF & F2F & FS & \multicolumn{1}{c|}{NT} & FF++  \\ 
\midrule
X-ray \cite{li2020face}      & 99.17           & 98.57          & 98.21          & \multicolumn{1}{c|}{98.13}          & 98.52          \\
PCL+I2G \cite{zhao2021learning}         & \textbf{100.00} & 98.97          & 99.86          & \multicolumn{1}{c|}{97.63}          & 99.11          \\
SBI \cite{shiohara2022detecting}    & 99.99           & \textbf{99.90} & 98.79          & \multicolumn{1}{c|}{98.20}          & 99.22          \\
AUNet \cite{bai2023aunet}           & 99.98           & 99.60          & \textbf{99.89} & \multicolumn{1}{c|}{\underline{98.38}}          & \underline{99.46}          \\ \midrule
KID (Ours)                    & \textbf{100.00} & \underline{99.87}          & \underline{99.81}          & \multicolumn{1}{c|}{\textbf{98.47}} & \textbf{99.59} \\ 
\bottomrule
\end{tabular}
\caption{Cross-manipulation evaluation on FF++ dataset. All forgery methods are unseen during training.}
\label{table:Cross manipulation}
\vspace{-10pt}
\end{table}

\begin{table}
\begin{tabular}{p{0.5cm}<{\centering}p{0.4cm}<{\centering}p{0.9cm}<{\centering}p{0.5cm}<{\centering}p{0.8cm}<{\centering}p{0.8cm}<{\centering}p{0.8cm}<{\centering}p{0.9cm}<{\centering}}
\toprule
\multirow{2}{*}{KIM} & 
\multirow{2}{*}{SBI} & 
\multirow{2}{*}{\begin{tabular}[c]{@{}c@{}}CGFLB\end{tabular}} & \multirow{2}{*}{\begin{tabular}[c]{@{}c@{}}S\&C\end{tabular}} & \multicolumn{4}{c}{Test Set AUC(\%)} \\
\cmidrule{5-8}
 & & & & FF++ & DFD & CDF & DFDC \\
\midrule

\usym{2717} & \usym{2717} & \usym{2717} & \usym{2717} & 96.36 & 82.78 & 73.49 & 61.55 \\

\usym{2713} & \usym{2717} & \usym{2717} & \usym{2717} & 98.93 & 96.77 & 87.35 & 70.94 \\

\usym{2713} & \usym{2713} & \usym{2717} & \usym{2717} & 96.11 & 98.06 & 89.72 & 72.03 \\

\usym{2713} & \usym{2713} & \usym{2713} & \usym{2717} & 98.84 & 98.04 & 91.49 & 74.05 \\

\usym{2713} & \usym{2713} & \usym{2713} & \usym{2713} & \textbf{99.59} & \textbf{99.46} & \textbf{95.74} & \textbf{75.77}\\
\bottomrule
\end{tabular}
\caption{The effect of key components of our KID. KIM (3.1), CGFLB (3.2) and S\&C (3.3) refer to the three key components of our method.}
\label{table:Ablation}
\end{table}

\cref{table:Cross manipulation} shows the performance of our method when tested on four manipulation methods present in the FF++ dataset. Since synthetic fake data is used during training, these four manipulation methods are unseen to the model. According to the results, our method achieves the best performance on the DF and NT forgery techniques, as well as overall. Moreover, the proposed method secures the second-best performance on F2F and FS methods, with only marginal differences of 0.03\% (F2F) and 0.08\% (FS) from the best results. These results demonstrate the excellent generalization capability of the proposed method across various manipulation methods.

\subsection{Ablation Study}

\noindent\textbf{Effect of Each Component.} We conduct quantitative experiments on the three key components of our method and the SBI \cite{chen2022self} method used in the training process to evaluate the contribution of each part.  The experimental results are presented in \cref{table:Ablation}. As shown in the table, it is evident that each of the three proposed components contributes significantly to the model's performance across all four test datasets. When all components are integrated, the model achieves the best overall performance. Specifically, the knowledge injection module (3.1) plays the most crucial role, improving the model's performance by an average of 9.95\% while the coarse-grained forgery localization branch (3.2) and the layer-wise suppression and contrast losses (3.3) also contribute significantly to overall effectiveness.
\vspace{5pt}

\begin{table}
  \centering
  \begin{tabular}{lcccc|c}
    \toprule
        \multirow{2}{*}{Method} & \multicolumn{5}{c}{Test Set AUC(\%)}\\ 
        \cmidrule{2-6}
        ~ & FF++ & DFD & CDF & DFDC & Avg \\ 
    \midrule
    ViT w/o KI & 96.36 & 82.78 & 73.49 & 61.55 & 78.55 \\    
    ViT w/ KI &  \textbf{99.59} & \textbf{99.46} & \textbf{95.74} & \textbf{77.57} & \textbf{93.09} \\
    \midrule
    DinoV2 w/o KI  & 98.66 & 96.87 & 78.83 & 65.95 & 85.08\\
    DinoV2 w/ KI & \textbf{99.73} & \textbf{99.31} & 
    \textbf{82.70} & \textbf{72.98} & \textbf{88.68} \\
    \midrule
    LeVit w/o KI  & 95.49 & 92.57 & 72.10 & 60.86 & 80.25 \\
    LeVit w/ KI & \textbf{97.20} & \textbf{93.16} & \textbf{79.75} & \textbf{67.25} & \textbf{84.34} \\ 
    \bottomrule
  \end{tabular}
  \caption{Generalization performances of different pre-trained transformer models with and without the knowledge inject framework.}
  \label{table:Different backbone}
\vspace{-10pt}
\end{table}

\begin{figure*}
    \centering
    \includegraphics[width=0.8\textwidth]{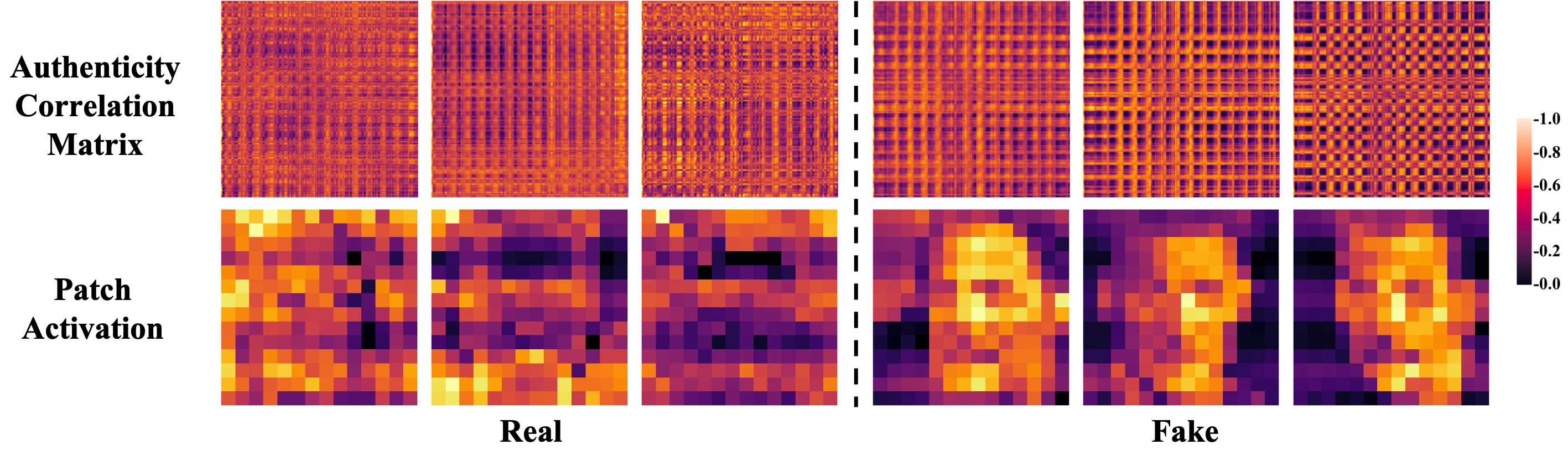}
    \caption{Visualization of the authenticity correlation matrix in \cref{eq:self-attention}. The Patch Activation in the second row represents the average correlation between each patch and all other patches in the matrix. }
    \vspace{-5pt}
    \label{fig:Knowledge}
    
\end{figure*}
\vspace{-5pt}

\begin{figure}
	\subfloat[Train Loss]{
	\begin{minipage}[b]{0.45\linewidth}
        {\includegraphics[scale=0.3]{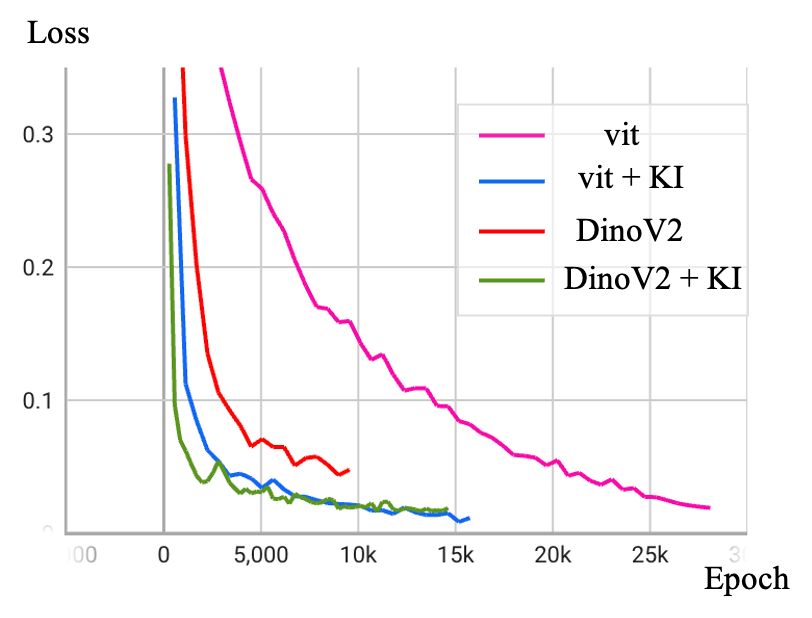}}
	\end{minipage}
}
	\subfloat[Validation AUC]{
	\begin{minipage}[b]{0.45\linewidth}
        {\includegraphics[scale=0.3]{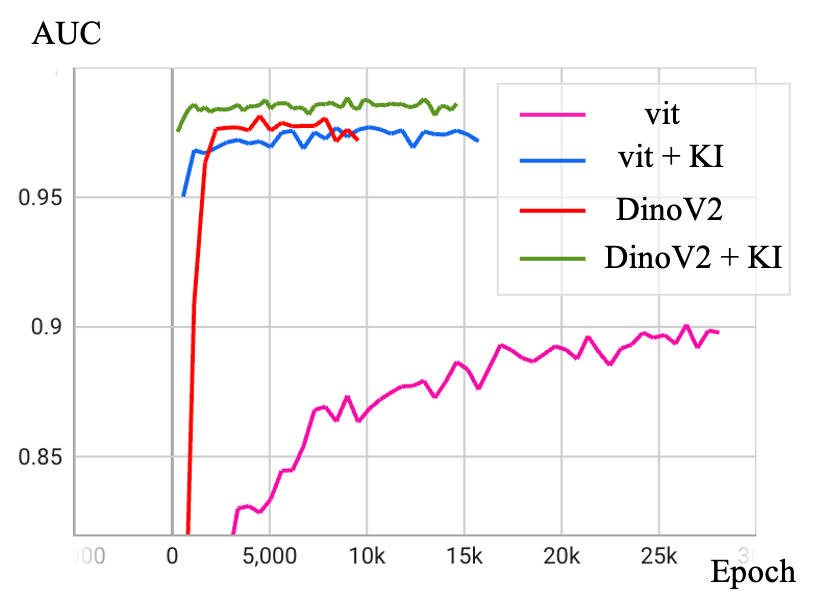}}
	\end{minipage}
}
\caption{Training loss and validation AUC curves of different pre-trained models after incorporating the knowledge injection framework.}
\label{fig:Injection Curve}
\end{figure}

\noindent\textbf{Impact on the Pre-trained Models.} We also apply the knowledge injection framework to the Vit-based backbone models DinoV2 \cite{oquab2023dinov2} and the more lightweight LeVit \cite{graham2021levit}. The experimental results, as shown in \cref{table:Different backbone}, demonstrate that the proposed framework knowledge injection framework possesses excellent compatibility with various Vit-based backbone models across multiple scales.

Furthermore, we analyze the impact of the knowledge injection framework on the training process, as illustrated in \cref{fig:Injection Curve}. With the knowledge injection framework incorporated, the model converges faster(4x faster for ViT, 2x faster for DinoV2) and exhibits better performance on both training and validation sets. This demonstrates that the knowledge injection framework effectively enhances model training efficiency and performance, highlighting its compatibility with ViT and DinoV2.

\begin{figure} [t]
	\subfloat[Vit]{
    	\begin{minipage}[b]{0.5\linewidth}
            {\includegraphics[scale=0.30]{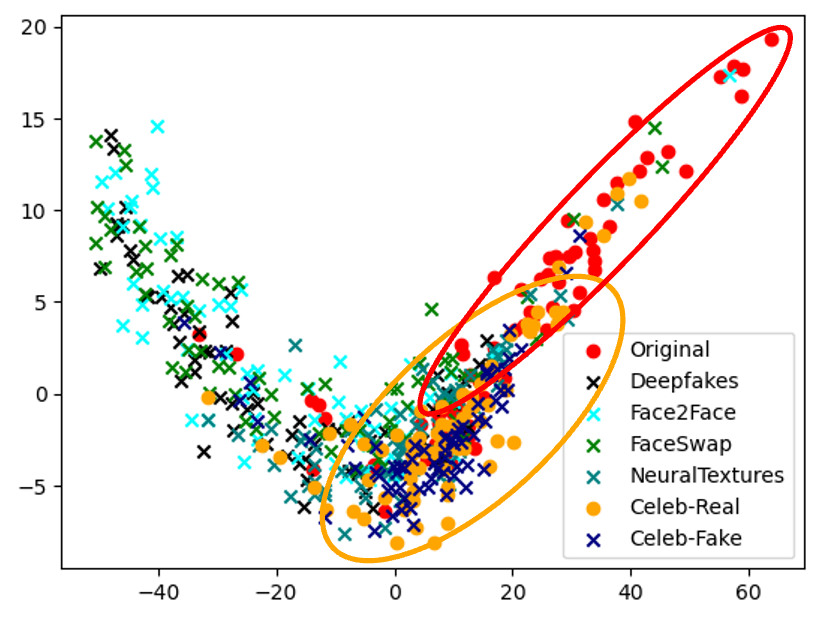}}
    	\end{minipage}
        }
	\subfloat[KID]{
	\begin{minipage}[b]{0.5\linewidth}
        {\includegraphics[scale=0.30]{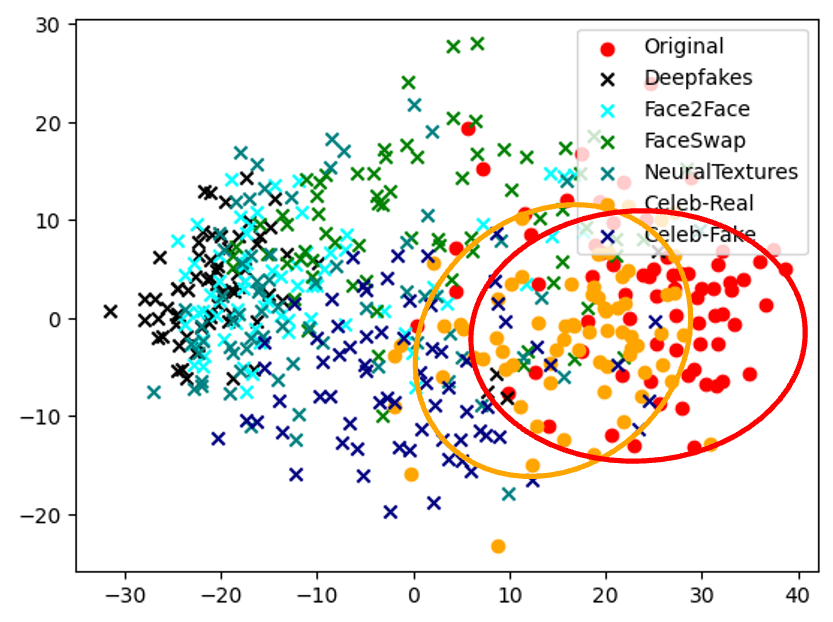}}
	\end{minipage}
}
\caption{The PCA feature spaces visualization of the basic ViT(a) and KID(b). The dots represent real images, the crosses represent forged images, and different colors represent different forgery methods in CDF and FF++ test sets. The circles represent the boundaries of real image features within and across domains. Best viewed in color.}
\label{fig:PCA}
\vspace{-10pt}
\end{figure}

\subsection{Qualitative Analysis}
\noindent\textbf{Knowledge Visualization.} In \cref{fig:Knowledge}, we visualize the learned knowledge as described in \cref{eq:self-attention}. The real images exhibit an average correlation due to the inherent consistency, while the fake images display significant periodic fluctuations, reflecting distinct regions with different distributions. We further obtain patch activation by averaging correlations across patches. The activation reveal high internal correlation in swapped areas of fake images and low correlation elsewhere, whereas real images show no such pattern This demonstrates that the learned knowledge effectively aids in distinguishing real from fake images.

\vspace{5pt}

\noindent\textbf{Feature Space.} We perform PCA \cite{abdi2010principal} dimensionality reduction on the final layer features of the test set images extracted by the model to observe the learned feature space, as shown in \cref{fig:PCA}. While the baseline model effectively differentiates real images within the training set, it struggles with distinguishing real images across domains from fake ones. In contrast, KID exhibits more accurate classification boundaries and more consistent distributions of real images across datasets, despite some overlap between high-quality forgeries and the real domain for CDF and NT. This highlights the improved representation of both real and fake images.
\section{Conclusion}

In this paper, we observe that existing deepfake detection methods usually overlook the learned knowledge of real data, which limits their generalization ability to handle the unseen real and fake images. To address this issue, we propose a simple yet novel method named Knowledge Injection based deepfake Detection (KID), by constructing a multi-task learning knowledge injection framework that is compatible with ViT-based backbone models. In the proposed framework, we propose a knowledge injection module to improve the model’s understandings of the distributions of real and fake data, via injecting specifically learned knowledge to the backbone network. We construct a coarse-grained forgery localization branch to enrich the forgery knowledge learned by the knowledge injection module. We propose two layer-wise suppression and contrast losses to balance the learned knowledge of real and fake data. Extensive experiments demonstrate that our KID achieves state-of-the-art generalization performances and faster convergence speed, and our framework is compatible with ViT, DinoV2 and LeVit.
{
    \small
    \bibliographystyle{ieeenat_fullname}
    \bibliography{main}

}

\clearpage
\appendix
\counterwithout{figure}{section} 
\counterwithout{table}{section}  
\renewcommand{\thefigure}{A\arabic{figure}} 
\renewcommand{\thetable}{A\arabic{table}}
\setcounter{figure}{0} 
\setcounter{table}{0}
\setcounter{page}{1}
\maketitlesupplementary

\section{Layer-wise Injection Analysis}
\begin{figure}
    \centering
    \includegraphics[width=\linewidth]{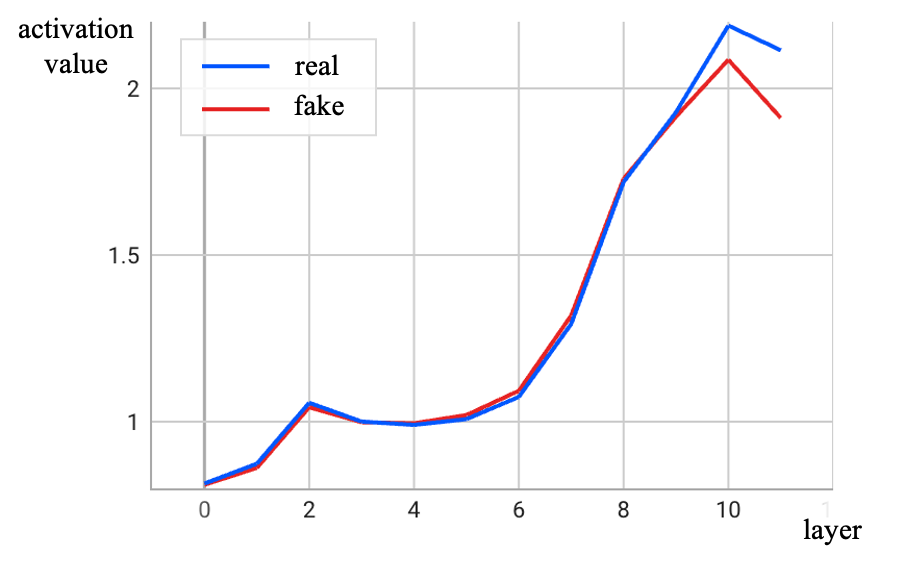}
    \caption{Layer-wise activation value of injection knowledge.}
    \label{fig:Activation Value}
\end{figure}

To explore the differences in the knowledge of real and fake images learned by the model, we computed the activation value of each layer’s authenticity correlation matrix in the model, following \cref{eq:activation value}. The results are displayed in \cref{fig:Activation Value}. Specifically, shallow layers inject less knowledge, preserving more fundamental representations of real data distributions. In contrast, deep layers learn more distinctions of the inconsistencies between real and fake data, resulting in more pronounced differences in the learned knowledge. These findings are consistent with the conclusions presented in \cref{subsec:Layer-Wise Contrast and Suppression}.

\begin{figure*}
    \centering
    \includegraphics[width=\linewidth]{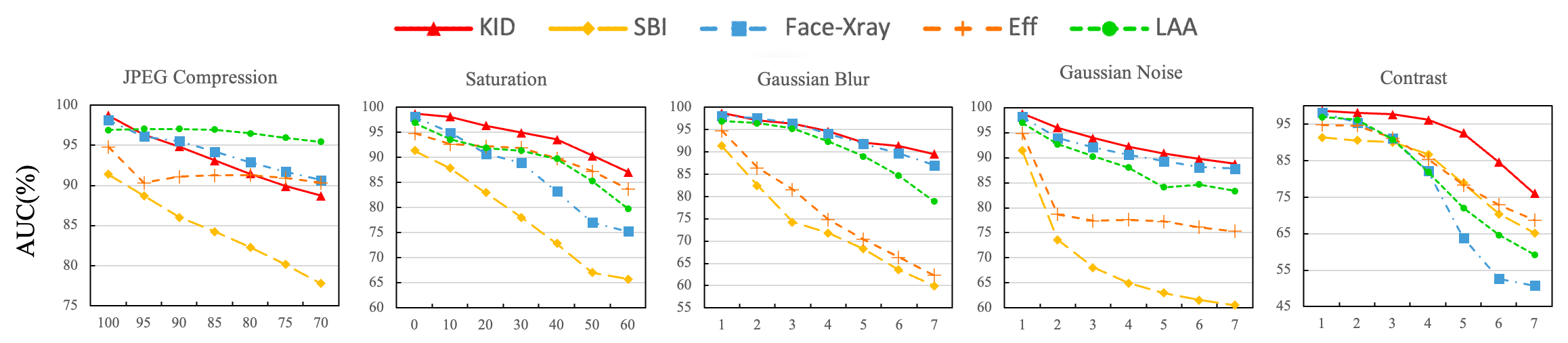}
    \caption{Robustness to different image degradation.}
    \label{fig:robust}
\end{figure*}

\begin{figure}

    \centering
    \includegraphics[width=0.82\linewidth]{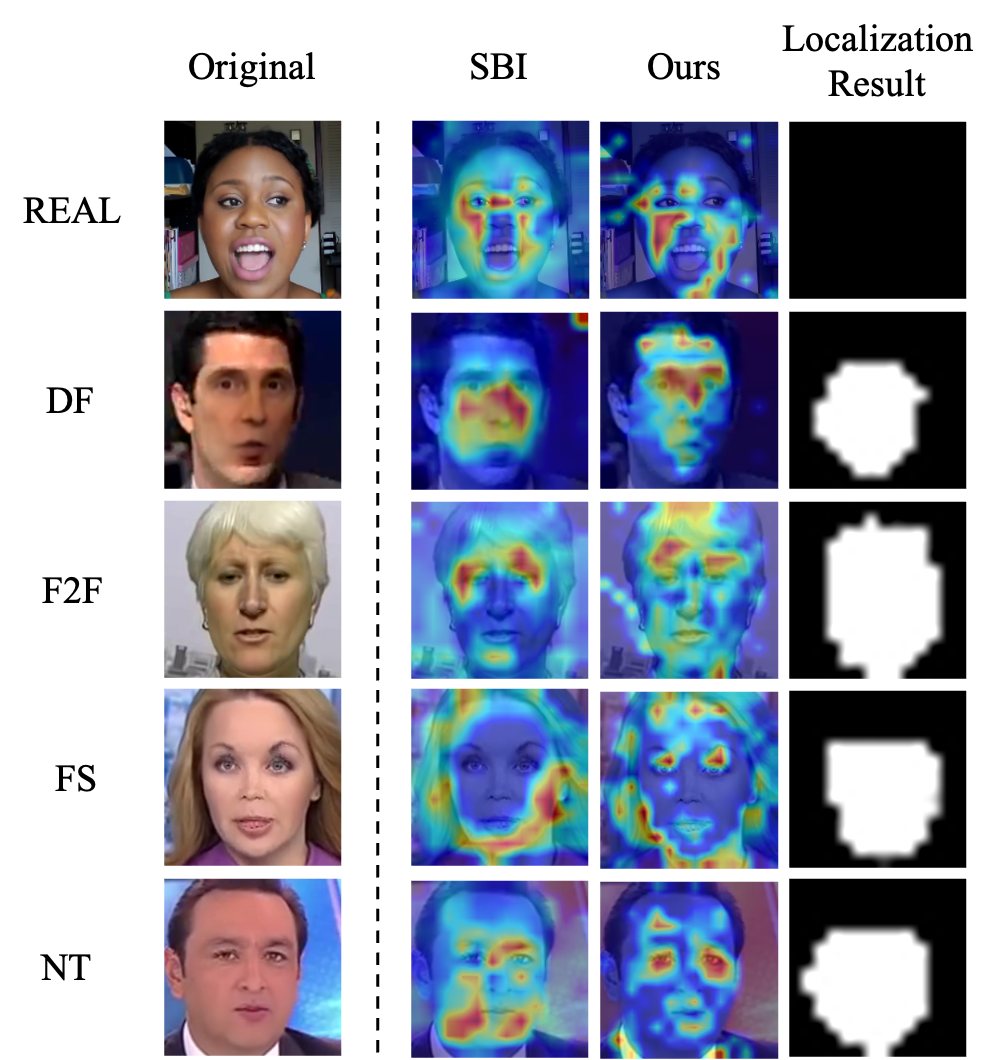}
    \caption{GradCAM \cite{selvaraju2017grad} visualization of images from different methods in the FF++ dataset and coarse-grained localization results.}
    \label{fig:CAM}
    
\end{figure}

\begin{table}
    \centering
    \begin{tabular}{c|ccc}
    \toprule

        \multirow{2}{*}{Methods} &  \multicolumn{3}{c}{Test Set AUC(\%)}  \\ \cmidrule{2-4}
         & DFD & CDF & FFIW \\ 
         \midrule
        Vit + LoRA \cite{hulora} 
        & 94.80 & 85.48 & 67.30 \\
        KID (Ours) & \textbf{99.46} & \textbf{95.74} & \textbf{82.53}  \\ \bottomrule
    \end{tabular}
    
    \caption{Comparison with LoRA.}
  \label{table:LoRA}
\end{table}

\section{Saliency Map.} In \cref{fig:CAM}, we present GradCAM \cite{selvaraju2017grad} visualization comparison between the SBI method and our proposed method, along with the coarse-grained localization results of our approach. Compared to CNN-based SBI methods, our approach places greater emphasis on key areas like the eyes and mouth, extracting richer information from more scattered and detailed facial regions. The localization results demonstrate that our coarse-grained forged localization branch can exploit knowledge to perform coarse localization tasks, allowing the knowledge injection module to learn more location information during training.

\begin{table}
    \centering
    \begin{tabular}{cc|ccc}
    \toprule
        \multirow{2}{*}{$\beta$} & \multirow{2}{*}{$\mu$} &  \multicolumn{3}{c}{Test Set AUC(\%)}  \\ \cmidrule{3-5}
         & & DFD & CDF & DFDC \\ 
         \midrule
         1.0 & 0.1 & \underline{99.00} & \textbf{95.86} & 72.87 \\
         1.5 & 0.1 & 98.37 & 93.53 & 70.66 \\
         1.2 & 0.05 & 97.45 & 93.99 & 71.12 \\
         1.2 & 0.2 & 98.80 & 95.46 & \underline{74.22} \\
         1.2 & 0.1 & \textbf{99.46} & \underline{95.74} & \textbf{75.77}  \\ \bottomrule
    \end{tabular}
    
    \caption{The impact of the hyper-parameters $\beta$ and $\mu$.}
    \label{tabel:hyper}
\end{table}

\section{Hyper-parameter Ablation Study}
 We conduct ablation experiments on the $\beta$ and $\mu$ hyperparameters in Layer-Wise Suppression and Contrast Losses to evaluate their impact on model performance. As shown in \cref{tabel:hyper}, the overly strict restrictions on shallow layers hinder the model's learning ability, reducing generalization performance, while excessively loose restrictions lead to overfitting. For contrast loss in deeper layers, a smaller limit $\mu$ diminishes the model’s ability to distinguish between real and fake images, while a higher limit $\mu$, though less deviating from optimal performance, causes the model to overemphasize the differences, making it harder to converge to an optimal solution. Based on the above experiments, we selected $\beta = 1.2$ and $\mu = 0.1$ for our final model configuration.

\section{KID v.s. LoRA}
Since both LoRA\cite{hulora} and the knowledge injection framework in this paper perform branch-based partial fine-tuning for downstream generalization, we compare our proposed KID method with LoRA in the context of deepfake detection. As shown in \cref{table:LoRA}, our method significantly outperforms LoRA in terms of generalization, demonstrating that our approach leverages knowledge more effectively through multiple learning strategies in deepfake detection tasks.

\section{Robustness Experiments}
To assess the robustness of our method, we compare the performance with other approaches under various image degradation scenarios. The experiment involves five common degradation strategies: JPEG compression, saturation change, Gaussian blur, Gaussian noise, and contrast change. As shown in \cref{fig:robust}, while all methods experience varying degrees of performance degradation as the image quality deteriorates, our method consistently maintains high detection accuracy. This demonstrates that our approach is resilient to different post-processing effects, making it well-suited for real-world forgery detection tasks.

\end{document}